\documentclass{bmvc2k}

\usepackage{arydshln}
\usepackage[dvipsnames]{xcolor}
\usepackage{multirow}
\usepackage{amssymb}
\usepackage{graphbox,graphicx}
\usepackage{soul}

\newcommand{\myparagraph}[1]{\textbf{#1}\ }
\usepackage{textcomp}
\title{Knowledge Distillation for End-to-End \\Person Search}

\addauthormulti{Bharti Munjal}{m.bharti@osram.com}{1}{2}
\addauthor{Fabio Galasso}{f.galasso@osram.com}{2}
\addauthor{Sikandar Amin}{s.amin@osram.com}{2}

\addinstitution{
 Technische Universit{\"a}t M{\"u}nchen,\\
 Boltzmannstra{\ss}e 3 Garching,\\
 Munich 85748, Germany
}

\addinstitution{
 OSRAM GmbH, \\ 
 Parkring 33 Garching,\\
 Munich 85748, Germany
}

\runninghead{B.MUNJAL ET. AL}{KNOWLEDGE DISTILLATION FOR
PERSON SEARCH}

\def\etal{\emph{et al}\bmvaOneDot}

\begin{document}

\maketitle

\begin{abstract}
We introduce knowledge distillation for end-to-end person search.
End-to-End methods are the current state-of-the-art for person search that solve both detection and re-identification jointly.
These approaches for joint optimization show their largest drop in performance due to a sub-optimal detector.

We propose two distinct approaches for extra supervision of end-to-end person search methods in a teacher-student setting. 
The first is adopted from state-of-the-art knowledge distillation in object detection. We employ this to supervise the detector of our person search model at various levels using a specialized detector. 
The second approach is new, simple and yet considerably more effective. This distills knowledge from a teacher re-identification technique via a pre-computed look-up table of ID features. It relaxes the learning of identification features and allows the student to focus on the detection task. This procedure not only helps fixing the sub-optimal detector training in the joint optimization and simultaneously improving the person search, but also closes the performance gap between the teacher and the student for model compression in this case. 
Overall, we demonstrate significant improvements for two recent state-of-the-art methods using our proposed knowledge distillation approach on two benchmark datasets.
Moreover, on the model compression task our approach brings the performance of smaller models on par with the larger models.
\end{abstract}
\section{Introduction}
\label{sec:intro}
Person search is the complex (multi-)task of jointly localizing people and verifying their identity against a provided query person ID. Person search has recently gained attention \cite{Liu2017NPSM,xiao2017joint,Xu2014PSS,zheng2016prw}, also thanks to its numerous applications, including cross-camera visual tracking, person verification, and surveillance.
The two tasks in person search, i.e.\ detection and re-identification are arguably in contrast with each other. In fact, detection should disregard any specific nuance of individuals and just retrieve any person, while re-identification should focus on these nuances, to distinguish individuals and retrieve the queried person. The dispute reflects in literature arguing that detection and re-identification in person search should be addressed separately~\cite{Xu2014PSS,zheng2016prw,Chen_2018_ECCV} or jointly~\cite{Liu2017NPSM,xiao2017joint,Xiao2017IANTI,munjal2019cvpr}.

\begin{figure}
\begin{center}
\subfigure{\includegraphics[align=c,width=4cm]{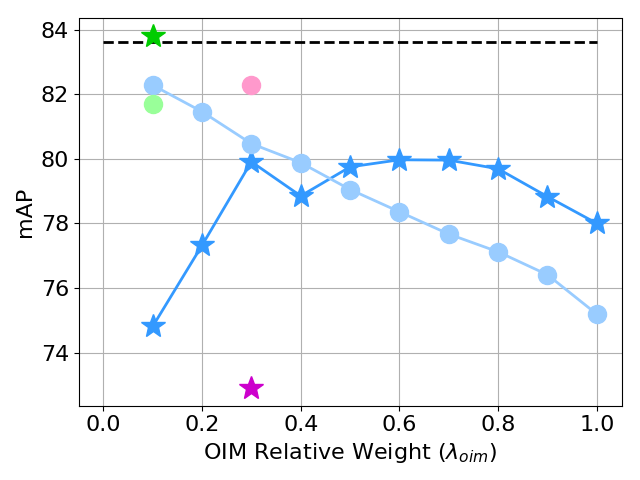}}
 \vspace{0pt} \subfigure{\includegraphics[align=c,width=4cm]{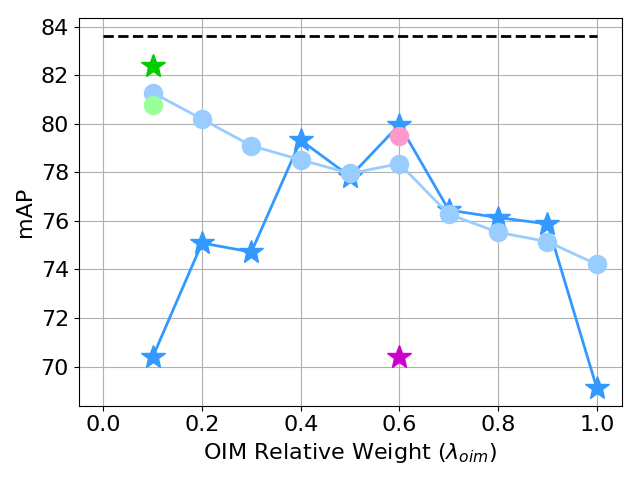}}
 \vspace{0pt} \subfigure{\includegraphics[align=c, width=4cm]{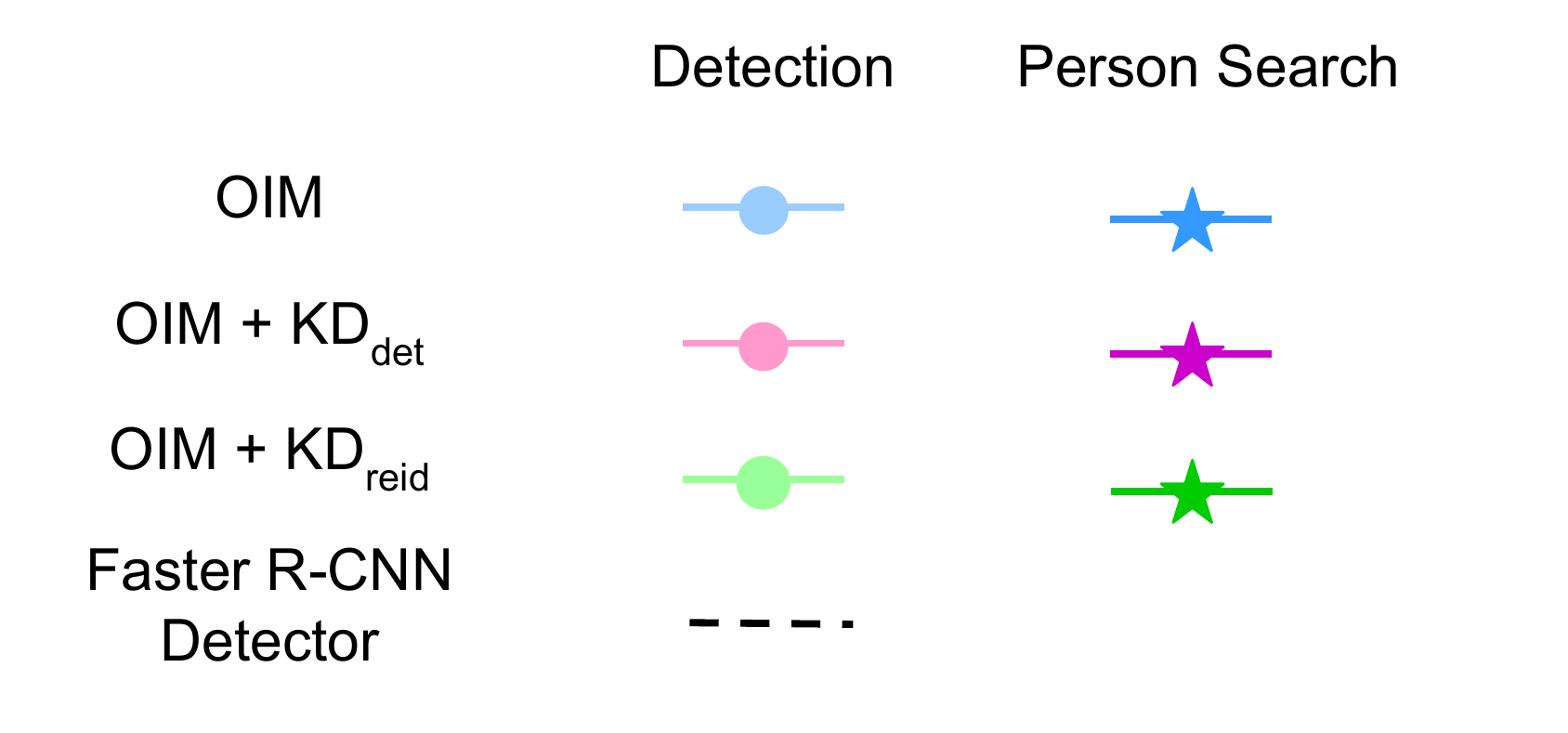}}
\caption{\footnotesize Joint detection and re-identification provides best performance~\cite{xiao2017joint,munjal2019cvpr} but analysis shows that detection is harmed and limits performance. Cyan (dot-marker) and blue (star-marker) curves illustrate the OIM~\cite{xiao2017joint} detection and person search performance respectively, when varying the relative training importance via the $\lambda_{oim}$ weight, cf.\ \eqref{eq:loss}. Both in the case of ResNet50 (\textit{left}) and ResNet18 (\textit{right}), the more weight is given to re-id task, the more the detector is harmed (decreasing cyan curve), which remains below the original Faster R-CNN performance (\textit{black dashed line}).
A teacher-student framework for the detection part of OIM, OIM + $\mathrm{KD_{det}}$ (\textit{magenta}), improves detection (light magenta, dot marker) but harms re-id, which is visible by the drop in person search performance (dark magenta, star marker).
The new knowledge distillation for the re-id part, OIM+$\mathrm{KD_{reid}}$ (\textit{green}) improves overall person search (dark green, star-marker) and also keeps the original detector performance.}
\label{fig:oim_weights}
\vspace{-1.0cm}
\end{center}
\end{figure}

Most recent state-of-the-art work in person search~\cite{xiao2017joint,munjal2019cvpr} demonstrates the benefits of end-to-end optimization. The approaches add the re-identification task onto the Faster R-CNN~\cite{ren2015faster} detection framework and learn both objectives jointly.
As a result, the detector performance degrades, as illustrated in Fig.~\ref{fig:oim_weights}, but it still remains state-of-the-art w.r.t person search performance.

In this work, we propose knowledge distillation to address the sub-optimal detector performance for end-to-end person search, and model compression to reach the same level of performance with a smaller model.
%
Knowledge distillation~\cite{NIPS2015_distilling} stems from the belief that training a neural network from labelled data requires a large amount of over-parameterization, as it is the case for the teacher, generally a large and accurate model. The teacher supervisory training signal enables training of a student~\cite{YangAAAI18teachers}, typically smaller.

Here we propose two distinct ways to distill knowledge and supervise the two joint tasks of person search, i.e.\ detection and re-identification.
We build upon the end-to-end approach of OIM~\cite{xiao2017joint}, the de-facto building block of all most recent person search approaches \cite{munjal2019cvpr,Xiao2017IANTI,Liu2017NPSM}. 
For the detection part, we adopt the teacher-student framework of \cite{NIPS2017_6676} which employs a multi-task loss, for intermediate feature guidance, region proposals, classification, and regression outputs.
For the re-identification part, we propose a new, simple and yet effective strategy that relieves the student from learning a look-up table (LUT) for the identities in the training set. Instead, we copy and fix a pre-computed teacher LUT, which relaxes the student task of identification feature learning.
We test our new distilled person search models on the two most recent CUHK-SYSU~\cite{xiao2017joint} and PRW-mini~\cite{zheng2016prw,munjal2019cvpr} datasets, extending both the baseline OIM~\cite{xiao2017joint} technique and the most recent query-based method QEEPS~\cite{munjal2019cvpr}. 
We demonstrate performance improvement in all cases. We achieve 85.0\% mAP on CUHK-SYSU and 39.7\% mAP on PRW-mini.
Notably, the same approach allows to train a smaller student network, realizing therefore model compression~\cite{bucilaKDD06}. In fact, we show that a ResNet18 student still provides 84.1\% mAP on CUHK-SYSU with only 46\% of parameters of the larger Resnet50 teacher.

We summarize our contributions as follows: \textbf{i.}\ we introduce knowledge distillation for person search and propose two distinct teacher-student frameworks, i.e.\ for the detector and for re-identification parts;
\textbf{ii.}\ we integrate our approach into the OIM~\cite{xiao2017joint} and QEEPS~\cite{munjal2019cvpr} person search models; \textbf{iii.}\ we show significant improvement over baseline methods on both the CUHK-SYSU~\cite{xiao2017joint} and PRW-mini~\cite{zheng2016prw,munjal2019cvpr} datasets; \textbf{iv.}\ we also show that our knowledge distillation approach enables model compression without drop in performance.

\vspace{-0.5cm}
\section{Related Work}
\vspace{-0.1cm}
\label{sec:related}
In this section, we first discuss the literature on the multiple tasks encompassed within person search. Then we review prior art on knowledge distillation and model compression.

\myparagraph{Person Detection.}
There is a large body of work on person detection, from methods based on hand-crafted features~\cite{Felzenszwalb2009ObjectDW,Dollr2014FastFP} to deep convolutional neural network (CNN) feature-learning methods~\cite{Girshick2015FastR,Girshick2014RichFH,Yang2015ConvolutionalCF}.
Best CNN detectors are either one-stage~\cite{Liu2016SSDSS,Redmon2017YOLO9000BF} or two-stage~\cite{ren2015faster,Girshick2015FastR,Girshick2014RichFH}. The latter select object proposals via a region proposal network and then classify those into persons vs background.
In line with OIM~\cite{xiao2017joint}, we adopt Faster R-CNN with a ResNet~\cite{he2016resnet} backbone, due to its robustness and flexibility.


 
 

\myparagraph{Person Re-Identification.}
Person re-identification is the task of classifying the same individuals as provided by a query sample, within a gallery of cropped, centered and aligned persons.
Earlier approaches have focused on manual feature design \cite{Wang2007ShapeAA,Gray2008ViewpointIP,Farenzena2010PersonRB,Zhao2013UnsupervisedSL} and metric learning \cite{Liao2015PersonRB,Zhao2017PersonRB,Kstinger2012LargeSM,Li2015MultiScaleLF,Liao2015EfficientPC,Paisitkriangkrai2015LearningTR,Ali_2018_ECCV}.
More recent re-identification approaches are based on CNNs~\cite{ahmed2015cvpr,li2014cvpr_deepreid,Yi2014DeepML,zhang2017alignedreid} and mainly concerned with the estimation of a ID-feature embedding spaces, either via Siamese networks and contrastive losses \cite{ahmed2015cvpr,li2014cvpr_deepreid,Liu2017EndtoEndCA,Varior2016ASL,Xu2018AttCompNet,Yi2014DeepML,Cheng2016PersonRB,Ding2015DeepFL,Chen2017BeyondTL}, or with ID-classification and cross-entropy losses \cite{Xiao2016LearningDF,Zheng2016MARSAV}.

\myparagraph{Person Search.}
Xu~\etal~\cite{Xu2014PSS} introduced this as finding a person in a set of non-cropped gallery images, given a crop of the queried person.
Person search involves detecting people in gallery images as well as verifying their ID against the provided query-ID.
It encompasses therefore the two tasks of detection and re-identification. Thanks to large-scale datasets (CUHK-SYSU~\cite{xiao2017joint} and PRW~\cite{zheng2016prw}), person search has witnessed progress but it remains divided into approaches addressing the two tasks separately~\cite{Xu2014PSS,zheng2016prw,Chen_2018_ECCV, person_eccv18} \emph{vs} jointly~\cite{Liu2017NPSM,xiao2017joint,Xiao2017IANTI,munjal2019cvpr,Yan_2019_CVPR}. 
We consider the two tasks jointly, since it was proven most recently beneficial~\cite{munjal2019cvpr}.

\myparagraph{End-to-End Person Search.}
Xiao~\etal~\cite{xiao2017joint} introduced a model for the end-to-end training of joint person search.
They extended Faster R-CNN to estimate an ID embedding for re-identification, and introduced an Online Instance Matching (OIM) loss, to effectively train it.
IAN \cite{Xiao2017IANTI} also proposed an end-to-end approach using a center loss~\cite{Wen2016ADF} with the goal to improve the intra-class feature compactness.
More recently, Munjal \etal~\cite{munjal2019cvpr} proposed query-guidance for OIM, dubbed QEEPS, i.e.\ conditioning person search on the non-cropped query image.
We propose knowledge distillation for the end-to-end OIM~\cite{xiao2017joint} and demonstrate the generality of our approach by also applying it to the current state-of-the-art QEEPS~\cite{munjal2019cvpr}.




\myparagraph{Knowledge Distillation and Model Compression.}
Knowledge distillation, proposed by \cite{bucilua2006model,NIPS2014_5484} and popularized by \cite{NIPS2015_distilling}, aims to train a small neural network from a large and more accurate one.
It has gained attention for its promise of more effective training~\cite{Romero2015-iclr}, better accuracy~\cite{belagiannisECCV2018} and efficiency~\cite{NIPS2014_5484,polino2018iclr,urban2016ArXiv,XuBMVC2018}, but it remains strongly limited to networks solving the single classification task.
When moving from classification to detection, as only few works~\cite{DBLP:journals/corr/ShenVBK16,NIPS2017_6676,Li_2017_CVPR} attempted, complex modelling questions arise as to where and how to supervise, which are not entirely answered yet. Such complex questions include the class importance imbalance, as for the background \emph{vs} the other classes, and the implicit multi-task objective, since detection implies the joint bounding box regression and classification.
Here, we apply knowledge distillation to the even more complex multi-task person search.

Distilling knowledge from a larger to a smaller network realizes model compression, i.e.\ train a smaller but accurate network. This has also been addressed for classification via quantization \cite{Zhou2017IncrementalNQ,Zhu2017TrainedTQ,Gong2014CompressingDC} and binarization \cite{Rastegari2016XNORNetIC,Binarization_NIPS2015_5647} of floating point operations, network pruning \cite{Carreira_2018_CVPR,Chen2015CompressingNN,Han2015LearningBW,Li2017PruningFF,Yu2018NISPPN} and factorization \cite{Sironi2013LearningSF,Lebedev2015SpeedingupCN}. 
The proposed knowledge distillation is directly applicable and realizes model compression for person search for the first time.

\section{Background - Online Instance Matching}
\label{sec:oim}
\begin{figure}[t] 
\begin{center}
	\includegraphics[trim=0cm 4cm 4cm 0cm, clip=false, width=0.8\linewidth]{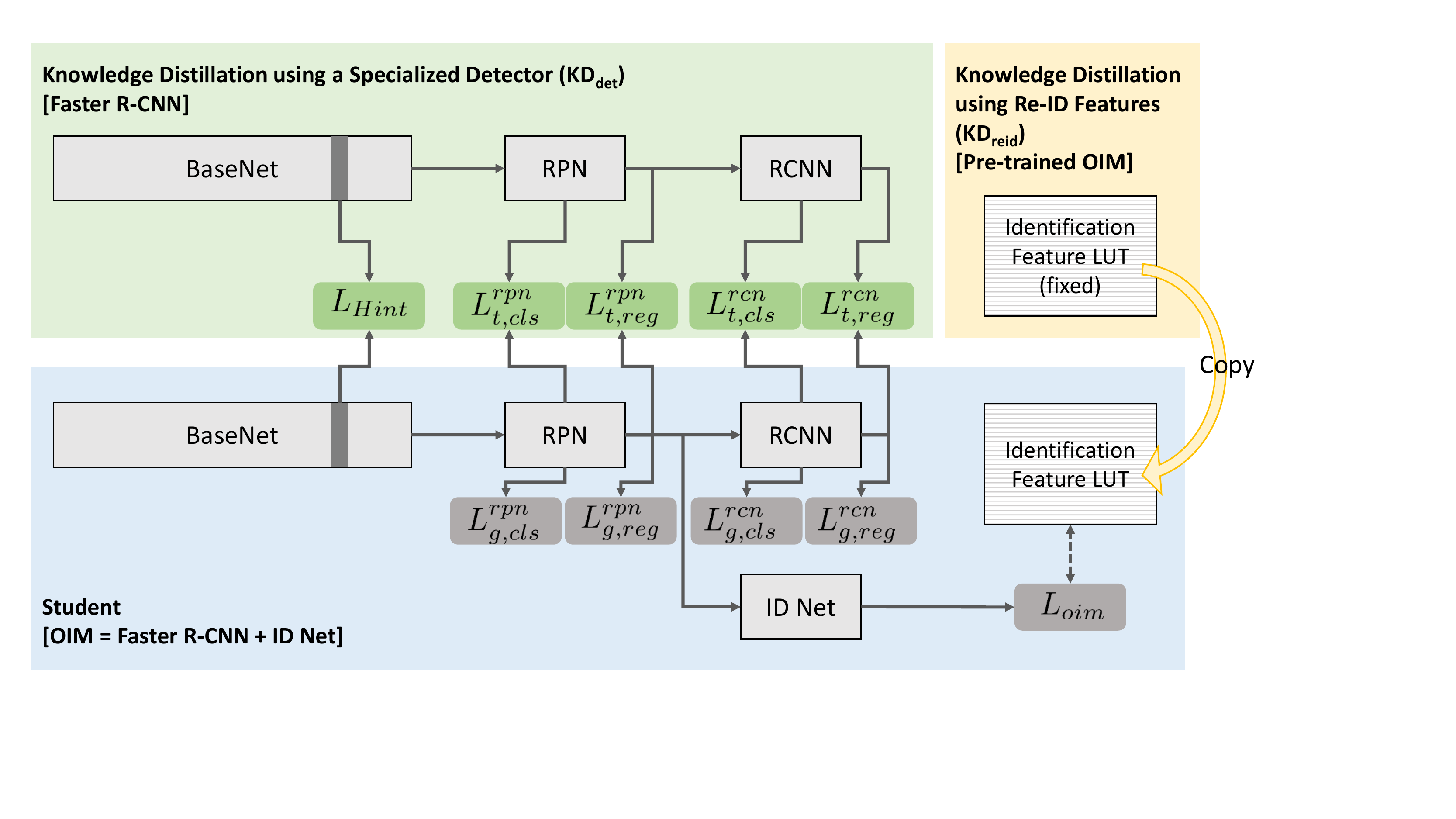}
	\caption{\footnotesize We propose two knowledge distillation approaches for person search. The first approach is motivated from ~\cite{NIPS2017_6676} and  uses the output of a specialized teacher detector (shown in green) to guide the detector of our person search student network (shown in blue). The second approach uses a copy of the LUT from a person search teacher (shown in yellow) and fix it during the student's training, thereby relaxing the task of ID feature learning and allowing the student to focus on the detection task.
	}\label{fig:network}
    \vspace{-0.8cm}
\end{center}
\end{figure}

Online Instance Matching (OIM) loss and the end-to-end architecture proposed by Xiao \etal~\cite{xiao2017joint} is currently the de-facto standard for identification feature learning in end-to-end person search \cite{Xiao2017IANTI,munjal2019cvpr,Liu2017NPSM}.
The architecture of \cite{xiao2017joint} is based on Faster R-CNN~\cite{ren2015faster} with a ResNet backbone~\cite{he2016resnet}, a Region Proposal Network (RPN) and a Region Classification Network (RCN), as illustrated in Fig.~\ref{fig:network} (blue region). 
In parallel to the RCN, \cite{xiao2017joint} defines an ID Net, which provides an identification feature embedding.
They introduce an OIM loss as an additional objective in the Faster R-CNN framework focusing on the task of learning unique identity (ID) features for the image instances of the same person.
This is accomplished by learning a lookup table (LUT) for all the identities in the training set. 
We refer to this approach as \textbf{OIM} in the text.

In more details, given ID features $x \in \mathbb{R}^{D}$ where $D=256$, OIM maintains a LUT $V \in \mathbb{R}^{D\mathrm{x}P}$ for D-dimensional ID features corresponding to the $P$ ID labels, and also a circular

queue $U \in \mathbb{R}^{D\mathrm{x}Q}$ containing most recent $Q$ ID features of the unlabeled identities that appear in the recent mini-batches.
During the forward pass of the network, the computed ID feature of each mini-batch sample is compared with all the entries in $V$ and $U$, to compute the OIM loss. The OIM objective is to minimize the negative expected log-likelihood. Given the softmax probabilities $p_t$ of the positive IDs  in the mini-batch, the OIM loss is given by $L_{oim} = -\mathrm{E}_x[\mathrm{log}\ p_t]$.

During the backward pass, entries of $V$ corresponding to the identities in the current mini-batch are updated using moving-average. The OIM detection and re-identification objectives are in conflict during the optimization, which results in a significantly sub-optimal detector performance, as illustrated in Fig.~\ref{fig:oim_weights}.
Intuitively, one would expect adjusting the relative weights of the two tasks might solve this problem. 
In fact we notice that the relative weights do play a role, and by decreasing the weight of the ReID task the detector approaches the performance of the specialized detector (standalone Faster R-CNN). However, at the same time the ReID task becomes harder to train and therefore its performance drops significantly.

\vspace{-0.4cm}
\section{Knowledge Distillation in OIM}
\label{sec:kd_oim}
We propose two independent approaches for knowledge distillation in OIM for person search. 
First, we draw from most recent literature in knowledge distillation for object detection to supervise the detector of the OIM model with the help of a specialized (hence stronger) person detector.
The second approach is to relax the task of re-identification with the help of a pre-computed identification feature table. As a result the detector becomes the focus during the optimization without compromising on the quality of identification features.
Both proposed approaches contribute in recovering the sub-optimal detector performance, as we would illustrate experimentally in Sec.~\ref{sec:experiments}.
However, we would demonstrate that the two approaches are not complementary.
In the following we discuss both approaches in detail.


\subsection{KD using a Specialized Detector}
\label{sec:kd_det}
As a first approach, we propose that the detector of the student model (OIM) to mimick the superior output of a stronger detector (a teacher). The teacher in this case is a specialized Faster R-CNN with the same backbone architecture as the student.
This approach stems from the belief that a better dedicated detector be a good teacher for the detector in OIM model, without modifying the OIM training.

To this end, we adopt the approach of Chen \etal~\cite{NIPS2017_6676} for the supervision of the student at various levels, from mimicking of the base features to the supervision of the detector. 
This corresponds to the green region in Fig.~\ref{fig:network}.
As illustrated in the figure, we employ supervision for three different components of the Faster R-CNN~\cite{ren2015faster} object detection framework.
Following \cite{NIPS2017_6676} these components are,
\textbf{i.} intermediate base feature representations using a hint based learning \cite{Romero2015-iclr},
\textbf{ii.} RPN classification and regression modules to produce better region proposals.
\textbf{iii.} likewise RCN classification and regression modules to generate stronger object detections.
We refer to this approach as $\bf{KD_{det}}$ in the text.


Following \cite{NIPS2017_6676,Romero2015-iclr}, hint loss ($L_{Hint}$) is given as an $\mathbb{L}_2$ loss between an intermediate layer of teacher BaseNet ($F_t$) and an adapted intermediate layer of student BaseNet ($f_a(F_s)$).
\begin{equation}
\begin{split}
&L_{Hint} = \| f_a(F_s) - F_t\| ^2_2 
\end{split}
\label{eq:losses}
\end{equation}

The classification loss ($L_{cls}$) and bounding box regression loss ($L_{reg}$) for both RPN and RCN sub-networks are given as:
\begin{equation}
\begin{split}
&L_{cls} = \mu L_{g,cls} + (1 - \mu) L_{t,cls} \\
&L_{reg} = L_{g,reg} + \gamma L_{t,reg}
\end{split}
\label{eq:losses}
\end{equation}




$L_{g,.}$ corresponds to the losses w.r.t the ground-truth labels \cite{ren2015faster} and $L_{t,.}$ corresponds to the losses w.r.t the output of the teacher network. Motivated from \cite{NIPS2017_6676}, we use soft cross entropy loss i.e $-\sum{P_t}\log{P_s}$ as $L_{t,cls}$ where $P_t$ and $P_s$ are the softened teacher and student classification probabilities with temperature 10. For $L_{t,reg}$ we use teacher bounded regression loss as in \cite{NIPS2017_6676}.
The overall optimization objective for this approach is given as follows:
\begin{equation}
\begin{split}
L &= L^{rpn}_{cls} + L^{rpn}_{reg} + L^{rcn}_{cls} + L^{rcn}_{reg} + \lambda_{Hint} L_{Hint} + \lambda_{oim} L_{oim}
\end{split}
\label{eq:loss}
\end{equation} 
whereby, $L_{oim}$ represents the OIM loss given in Sec.~\ref{sec:oim}.
For details on the contribution of each loss term (except $L_{oim}$), we refer the reader to \cite{NIPS2017_6676}. We set $\mu$, $\gamma$ and $\lambda_{Hint}$ to 0.5 as in \cite{NIPS2017_6676}, while we investigate different values of $\lambda_{oim}$ in Sec.~\ref{sec:experiments} and also Fig.~\ref{fig:oim_weights}. 




\vspace{-0.3cm}
\subsection{KD using Pre-Trained Re-ID Features}
\label{sec:kd_reid}
The OIM optimization includes an iterative estimation of \textbf{i.\ } the ID feature embeddings, and \textbf{ii.\ } the lookup table (LUT) for the evaluation of these ID feature embeddings and computation of the OIM loss.
The LUT in the original model \cite{xiao2017joint} is randomly initialized. It eventually converges over time, however this iterative complexity impacts the learning of the parallel detection task.

We propose a new approach to person search with OIM, whereby we leverage knowledge distillation to relax one of the two tasks, i.e. estimating the LUT. In other words, distilling knowledge for the re-identification means that the student is not tasked any more with training for the OIM LUT. Most importantly, originally an optimization goal becomes instead a supervisory signal which eases the training and improves the performance.

To accomplish this, we fix the LUT $V$ of our student OIM model using a copy from a pre-trained OIM model.
This approach for knowledge distillation is illustrated by yellow components in Fig.~\ref{fig:network}. We refer to this approach as $\bf{KD_{reid}}$ in the text.
%
The pre-computed $V$ is fixed, hence not updated during the back-propagation step. The network is able to obtain the optimal supervision for the ID features directly from the very first iteration.

The proposed $\mathrm{KD_{reid}}$ approach is in contrast with the $\mathrm{KD_{det}}$ approach, since it aims to simplify the learning of the ID features while the latter focuses on directly supervising and improving the detector.
Moreover, the proposed method do not add any overhead to the network training as compared to the original OIM. In fact it reduces the computations (FLOPS), since LUT $V$ is fixed and we skip the step of updating it during back-propagation.
Whereas, for $\mathrm{KD_{det}}$ a forward pass over teacher network is also required.

Finally, we would like to emphasize that the proposed knowledge distillation $\mathrm{KD_{reid}}$ is applicable to any OIM based person search method, since we just require to copy and fix the LUT $V$ from a pre-trained model.
We show this by applying $\mathrm{KD_{reid}}$ to the most recent work Query-guided End-to-End Person Search (QEEPS) \cite{munjal2019cvpr} in Sec.~\ref{sec:experiments}.


\section{Model Compression}


In this work, we also demonstrate achieving model compression using our proposed knowledge distillation approaches.
We do not require any modification to the proposed approaches which we discussed in section \ref{sec:kd_det} and \ref{sec:kd_reid}.
The only difference is that for model compression the backbone architecture of our student OIM network (blue region in Fig.~\ref{fig:network}) is much smaller than the specialized detector (green region), and the pre-trained OIM model (in yellow).

Prior works in knowledge distillation \cite{Denil_NIPS2013_5025,NIPS2017_6676} show that neural networks are often over-parametrized, and a proper teacher-student knowledge distillation has the potential to scale down the redundancy while keeping the performance intact. 
In other words, supervision from a stronger model as a teacher allows a weaker model to reach the level of performance which the smaller model itself, with current training procedures, is unable to arrive at.

In our evaluation we focus on different sizes of the Resnet~\cite{he2016resnet} architecture to study model compression. 
In particular, our teacher has a larger backbone architecture (Resnet50), while our student is based on Resnet18, which is the smallest standard variant of this architecture.
In Sec.~\ref{sec:experiments}, we discuss this further.






\section{Experiments}
\label{sec:experiments}


\noindent \myparagraph{CUHK-SYSU.} 
The CUHK-SYSU is a large scale person search dataset \cite{xiao2017joint} consisting of 18,184 images, labeled with 8,432 identities and 96,143 pedestrian bounding boxes (23,430 boxes are ID labeled). 
The dataset displays a large variation in person appearance, backgrounds, illumination conditions etc.
For our evaluations, we adopt the standard train/test split as detailed in \cite{xiao2017joint,munjal2019cvpr}.

\noindent \myparagraph{PRW-mini.}
PRW \cite{zheng2016prw} is another important dataset focusing on the task of person search. Different from CUHK-SYSU, PRW is acquired in a single setup, i.e. a university campus using six cameras. Overall it consists of 11,816 images with 34,304 bounding boxes and 932 identities.
The diversity in the background and the appearance of the persons is limited compared to CUHK-SYSU. 
Munjal \etal~\cite{munjal2019cvpr} proposed PRW-mini\footnote{\scriptsize Publicly available at \url{https://github.com/munjalbharti/Query-guided-End-to-End-Person-Search}}, a subset of PRW, for a faster and representative benchmarking of the full dataset. The motivation came from the huge computational complexity of their query-guided person search method (QEEPS), which we also evaluate in this work.
%

\noindent \myparagraph{Evaluation Metrics.}
Following \cite{xiao2017joint,Xiao2017IANTI,Liu2017NPSM,munjal2019cvpr}, we adopt mean Average Precision (mAP) and Common Matching Characteristic (CMC top-K) for results on person search. 
On the other hand, we report mAP for person detection results.
mAP metric is common in the detection literature, reflecting both precision and recall of results.
CMC is specific to re-identification and reports the probability of retrieving at least one correct ID within the top-K predictions (CMC top-1 is adopted here, which we refer to as top-1).

\noindent \myparagraph{Implementation Details.}\label{sec:impldetails}
Our implementation of OIM is based on Resnet \cite{he2016resnet} and uses first four blocks (conv1 to conv4) as BaseNet. The input images are re-scaled such that their shorter side is 600 pixels, unless specified otherwise. 
We employ the same training strategy as in \cite{munjal2019cvpr}. Note that, we augment the data by flipping and initialize the backbone architecture using pre-trained ImageNet~\cite{imagenet_cvpr09} weights.


\subsection{Knowledge Distillation in OIM}
In Table~\ref{tab:res50ablation_experiments}, we summarize the ablation study of our proposed knowledge distillation approaches, i.e. $\mathrm{KD_{det}}$ and $\mathrm{KD_{reid}}$, on CUHK-SYSU~\cite{xiao2017joint} dataset.
All experiments in this section consider Renset50 as the backbone architecture.
We begin with the evaluation of a pure detector (Faster R-CNN~\cite{ren2015faster}). We obtain a mAP of 83.6\% and recall of 88.1\% as the baseline for detection accuracy. We refer to this as DET$_{50}$ in the text.
Then, we evaluate a basic OIM \cite{xiao2017joint} model with a relative weight of ReID task $\lambda_{oim}=1$. We set this as our baseline for person search for rest of the experiments in this section and refer it as OIM$_{50}$, which achieves 78\% mAP and 77.9\% CMC top-1 for person search, meanwhile significantly underperforms on detector accuracy compared to DET$_{50}$ by 8.4\% mAP.
In Fig.~\ref{fig:oim_weights}, we investigate the effect of relative weighting of the detection and re-identification tasks by varying $\lambda_{oim}$ demonstrating contrasting nature of both tasks.

Next, we detail the results for our distillation approaches.
We show that applying $\mathrm{KD_{det}}$ to OIM with DET$_{50}$ as the teacher and $\lambda_{oim}=1$ does not change the person search performance, however reasonably improves the detector accuracy by 3.7pp mAP and 2.3pp recall.
On the other hand, selecting $\lambda_{oim}=0.3$ degrades the person search by more than 5\% in terms of mAP as well as top-1, but the detector recovers almost entire performance of teacher DET$_{50}$. This result is intuitive since the detector of the student model is getting an additional signal from the teacher network and at the same time we decreased the relative weight of the contrasting re-id task.

Keeping the results of $\mathrm{KD_{det}}$ in view, for our second distillation approach $\mathrm{KD_{reid}}$, we select $\lambda_{oim}=0.1$ to ensure the detector of the student gets the required focus during the joint optimization. Whereas, supervision through $\mathrm{KD_{reid}}$ is supposed to simplify the re-id task, hence justifying the lower value of its relative weight $\lambda_{oim}$.
Interestingly, we observe significant improvements for both detection (6.7pp mAP and 4.2pp recall) and person search (3.2pp mAP and 3.1pp top-1) over the baseline OIM$_{50}$. Note that, $\mathrm{KD_{reid}}$ is significantly simplified knowledge distillation approach in comparison to $\mathrm{KD_{det}}$, and still it improves significantly on both detection and re-id tasks unlike $\mathrm{KD_{det}}$.
This result further motivates the importance of research into appropriate supervision and optimal training procedures.

We further combine $\mathrm{KD_{det}}$ with $\mathrm{KD_{reid}}$ in our evaluations in Table~\ref{tab:res50ablation_experiments}. We notice that the detector of the student reached the Faster R-CNN detector performance (83.2\% mAP and 87.8\% recall), however there is a drop in person search results as compared to employing only $\mathrm{KD_{reid}}$ (80.3 vs 81.2 in mAP). Clearly, the addition of $\mathrm{KD_{det}}$ increases the focus on the detection task; weakening the training of the re-id branch, hence declining the person search results.
This result indicates the challenges in combining the two knowledge distillation approaches ($\mathrm{KD_{det}}$, $\mathrm{KD_{reid}}$). One could adjust the relative weighting of these approaches, we however adopt $\mathrm{KD_{reid}}$ for all of our next experiments, due to its superior performance and simplicity. \\


\tabcolsep 4pt

\begin{table}[t]
\scriptsize
\begin{center}
\begin{tabular}{lccccccccc}
 \hline
 \textbf{Student}&$\bf{\lambda_{oim}}$ & \textbf{Type of KD} & \textbf{Teacher} &\multicolumn{2}{c}{\textbf{Person Search}} & \multicolumn{2}{c}{\textbf{Detection}}\\
 Resnet50 Models &   &  & & mAP(\%) & top-1(\%) & mAP(\%) &  Recall(\%)\\
 \hline
 \hline
Faster R-CNN (DET$_{50}$)*  &-&- & - &-  & - & \textbf{83.6} & \textbf{88.1}\\
OIM  (OIM$_{50}$, Baseline)* &1.0 & -  & -   & 78.0  & 77.9 & 75.2 & 82.7\\
OIM &1.0 & $\mathrm{KD_{det}}$  & DET$_{50}$  & 78.3 & 77.5 & 78.9  & 85.0  \\

OIM  & 0.3 &$\mathrm{KD_{det}}$& DET$_{50}$ & 72.9 & 72.0 & 82.3  & 87.2 \\


OIM   & 0.1 &$\mathrm{KD_{reid}}$ & OIM$_{50}$  & 81.2 & 81.0 & 81.9 & 86.9 \\
OIM & 0.1 & $\mathrm{KD_{det}}$, $\mathrm{KD_{reid}}$ & DET$_{50}$, OIM$_{50}$  &  80.3 & 79.9 & 83.2 & 87.8 &\\
\hdashline[3pt/5pt]
OIM  & 0.1 & $\mathrm{KD_{reid}}$ &  OIM$_{18}$   &  75.0 & 73.9 & 81.7 & 86.8\\
OIM & 0.1 & $\mathrm{KD_{reid}}$ &  QEEPS$_{50}$   & 83.8& 84.2 & 81.7 & 86.8\\
QEEPS\cite{munjal2019cvpr} (QEEPS$_{50}$)*& 1.0 & -& -  & 84.4 &84.4 & - & -\\
QEEPS & 0.1 & $\mathrm{KD_{reid}}$ & OIM$_{50}$   &  83.2 &82.9& - & - \\
QEEPS & 0.1 & $\mathrm{KD_{reid}}$ & QEEPS$_{50}$  & \textbf{85.0}& \textbf{85.5} & - & - \\

\hline
\end{tabular}
\end{center}
\caption{\footnotesize Knowledge distillation for Resnet50 student models. Above the dashed line, are the detection and person search results of the student model using the two proposed distillation methods ($\mathrm{KD_{det}}$ and $\mathrm{KD_{reid}}$). For $\mathrm{KD_{det}}$, Resnet50 Faster R-CNN detector (DET$_{50}$) is used as teacher. For $\mathrm{KD_{reid}}$ Resnet50 OIM model (OIM$_{50}$) is used as teacher. Below the dashed line, are the results of using  different teachers with OIM and QEEPS student. OIM$_{18}$ represents Resnet18 OIM model, QEEPS$_{50}$ represents Resnet50 QEEPS model. (*) indicates models trained without KD.}
\label{tab:res50ablation_experiments}
\end{table}

\noindent \myparagraph{Significance of Teachers' Quality.}
In Table \ref{tab:res50ablation_experiments} below the dashed line, we study the effect of using different teacher models in $\mathrm{KD_{reid}}$, namely Resnet18 (referred as OIM$_{18}$) and recently proposed QEEPS \cite{munjal2019cvpr} with Resnet50 (referred as QEEPS$_{50}$), in addition to the baseline OIM$_{50}$. Notice, how using OIM$_{18}$ as teacher drops the person search performance to 75\% mAP and 73.9\% top-1 which is 3pp mAP and 4pp top-1 lower than the baseline OIM$_{50}$. While, using QEEPS$_{50}$ as teacher gives significant improvement in person search over the baseline, 5.8pp mAP and 6.3pp top-1. This is 2.6pp mAP and 3.2pp top-1 better than using OIM$_{50}$ as a teacher. Overall, we can conclude that stronger teachers provide stronger supervision hence better results, while inferior teachers may harm the performance. 
However, we also demonstrate that students using same model as teacher, e.g. OIM Resnet50 student, OIM$_{50}$ teacher (cf.~Table~\ref{tab:res50ablation_experiments}, $5^{th}$ row) also improves the results due to improved training conditions.
It is worth noting that the detector performance, in this case, remains almost the same when using different teachers for $\mathrm{KD_{reid}}$ and $\lambda_{oim}=0.1$ (mAP 81.9\%, 81.7\% and 81.7\% for OIM$_{50}$, OIM$_{18}$ and QEEPS$_{50}$, respectively), since a higher relative importance of the detector encourages its convergence to the performance of a pure detector (mAP 83.6\%).

Next, we also evaluate the state-of-the-art QEEPS model~\cite{munjal2019cvpr} with Resnet50 as a student with supervision of OIM$_{50}$ and QEEPS$_{50}$ as teachers. QEEPS does not consider intermediate detection stage, therefore we only report person search results. As shown in the table, using OIM$_{50}$ to train QEEPS dropped the person search performance by 1.2pp mAP and 1.5pp top-1, while using QEEPS$_{50}$, results in an improvement of 0.6pp mAP and 1.1pp top-1.\\

\tabcolsep 7pt

\begin{table}[t]
\begin{center}
\begin{tabular}{cc}
\scriptsize
\begin{tabular}[b]{lccc}

\hline
 Method & mAP(\%) & top-1 (\%) \\
\hline
\hline
 OIM \cite{xiao2017joint}  & 75.5 & 78.7 \\
 Distilled OIM & \textbf{83.8} & \textbf{84.2} \\
\hdashline[3pt/5pt]
QEEPS \cite{munjal2019cvpr}  & 84.4 & 84.4 \\
  Distilled QEEPS & \textbf{85.0} & \textbf{85.5} \\
  \hline
 IAN \cite{Xiao2017IANTI} & 76.3 & 80.1  \\
 NPSM \cite{Liu2017NPSM} & 77.9 & 81.2 \\
 Context Graph \cite{Yan_2019_CVPR} & 84.1 & 86.5\\
 CLSA \cite{person_eccv18} & 
87.2 & 88.5

\end{tabular}
&
\scriptsize
\begin{tabular}[b]{lcc}
\hline

Method & mAP(\%) & top-1 (\%) \\
\hline
\hline

OIM\ddag \cite{munjal2019cvpr} & 38.3 & 70.0 \\
Distilled OIM\ddag & \textbf{39.5} & \textbf{73.3} \\
\hdashline[3pt/5pt]
QEEPS \cite{munjal2019cvpr} & 39.1 & \textbf{80.0} \\
Distilled QEEPS & \textbf{39.7} & \textbf{80.0}\\
\hline
Mask-G \cite{Chen_2018_ECCV} & 33.1 & 70.0 
\end{tabular}\\
\hline
\footnotesize (a) CUHK-SYSU & \footnotesize  (b) PRW-mini

\end{tabular}
\vspace{-0.2cm}
\end{center}
\caption{\footnotesize Comparison to the state-of-the-art on, (a) CUHK-SYSU \cite{xiao2017joint} (image size = 600), and (b) PRW-mini \cite{munjal2019cvpr} (image size = 900), where OIM\ddag\ is same as in \cite{munjal2019cvpr}.}
\label{tab:sota}
\end{table}

\noindent \myparagraph{Comparison to the State-of-the-Art.} 
It is important to note that our knowledge distillation approach $\mathrm{KD_{reid}}$ can be directly applied to all methods that use OIM loss~\cite{xiao2017joint} for learning the identification features \cite{Yan_2019_CVPR,Chen_2018_ECCV,munjal2019cvpr,xiao2017joint}.
In Table \ref{tab:sota} (a), we demonstrate our approach on CUHK-SYSU~\cite{xiao2017joint} dataset for two such state-of-the-art methods, i.e. OIM~\cite{xiao2017joint} and QEEPS~\cite{munjal2019cvpr}.
Our distilled OIM (above the dashed line) using $\mathrm{KD_{reid}}$ from QEEPS$_{50}$ outperforms OIM~\cite{xiao2017joint} by 8.3pp in mAP and 5.5pp in top-1.
Similarly, our distilled QEEPS outperforms QEEPS~\cite{munjal2019cvpr} by 0.6pp mAP and 1.1pp top-1, achieving overall 85.0\% mAP and 85.5\% top-1. 
Also notice that our distilled QEEPS outperforms IAN~\cite{Xiao2017IANTI}, NPSM~\cite{Liu2017NPSM} and a recent context graph based approach~\cite{Yan_2019_CVPR} by 8.7pp, 7.1pp and 0.9pp mAP, respectively. 

Our knowledge distillation approach is also applicable to other methods like IAN~\cite{Xiao2017IANTI}, NPSM~\cite{Liu2017NPSM} and CLSA~\cite{person_eccv18} that learn the identification features using softmax loss.
In this case, first we would need to compute an OIM-like ID feature table for the teacher model and then use it to additionally supervise the identification feature learning of the student model.

%

In Table \ref{tab:sota} (b), we report results on PRW-mini. 
In this case we adopt image size of 900 pixels, following ~\cite{munjal2019cvpr,Shen_2018_ECCV}. Our distilled OIM\ddag\footnote{OIM\ddag\ uses image size 900 whereas OIM uses image size 600.} (above the dashed line) trained with $\mathrm{KD_{reid}}$ using QEEPS$_{50}$ surpasses OIM\ddag~ by 1.2pp mAP and 3.3pp top-1. 
Similarly our distilled QEEPS surpasses QEEPS~\cite{munjal2019cvpr} by .6pp mAP.  



\vspace{0.2cm}
\subsection{Model Compression}

In Table~\ref{tab:res18ablation_experiments}, we report the model compression results for our two knowledge distillation methods, $\mathrm{KD_{det}}$ and $\mathrm{KD_{reid}}$. We employ Resnet18 (with \textasciitilde 46\% of Resnet50 parameters), as the network architecture for all the student entries in this table. We keep DET$_{50}$ as the teacher for distillation using $\mathrm{KD_{det}}$.

The person search results for our baseline model OIM$_{18}$ i.e. 69.1\% mAP and 68\% top-1 are significantly lower than OIM$_{50}$ by around 9\%.
Whereas, the detector of OIM$_{18}$ is only slightly worse than OIM$_{50}$ but significantly below the pure detector accuracy of DET$_{18}$ (mAP 82.4\%).
As shown in Table \ref{tab:res18ablation_experiments}, the application of $\mathrm{KD_{det}}$ to OIM at $\lambda_{oim}=1$, improves the detection performance by 3.3pp mAP and 2.1pp recall with no effect on person search results. 
We also evaluate the $\mathrm{KD_{det}}$ at $\lambda_{oim}=0.6$, as it provides the best trade-off between detection and re-id tasks (cf.~Fig.~\ref{fig:oim_weights} (right)).
We notice that the detection improves by 1.3pp mAP and 1pp recall, while person search performance drops by a large value of 9.5pp mAP and 11.4pp top-1. This result indicates that while detector is being directly supervised through $\mathrm{KD_{det}}$, a lower $\lambda_{oim}$ for re-id task during optimization is counter-productive for person search.
\tabcolsep 4pt
\begin{table}[t]
\scriptsize
\begin{center}
\begin{tabular}{lccccccccc}
 \hline
 \textbf{Student}&$\bf{\lambda_{oim}}$ & \textbf{Type of KD} & \textbf{Teacher} &\multicolumn{2}{c}{\textbf{Person Search}} & \multicolumn{2}{c}{\textbf{Detection}}\\
 Resnet18 Models &   &  & & mAP(\%) & top-1(\%) & mAP(\%) &  Recall(\%)\\
 \hline
 \hline
Faster R-CNN (DET$_{18}$)* & - & - & -  & -  & - & \textbf{82.4} & \textbf{87.2}\\
OIM (OIM$_{18}$, Baseline)* & 1 & -  & -  & 69.1  & 68.0 & 74.2 & 81.6\\
OIM & 1 & $\mathrm{KD_{det}}$  & DET$_{50}$  & 69.1& 68.0&77.5 & 83.7 &
 & \\
OIM*    & 0.6 & - & - & 79.9  & 80.4 & 78.4 & 84.4\\
OIM  &0.6  & $\mathrm{KD_{det}}$ & DET$_{50}$  &70.4& 69.0&79.7&85.4 &\\


OIM  & 0.1 & $\mathrm{KD_{reid}}$ &  OIM$_{50}$ & {80.5} & {80.9} & 80.6 & 85.9\\
OIM & 0.1 & $\mathrm{KD_{det}}$, $\mathrm{KD_{reid}}$ & DET$_{50}$, OIM$_{50}$  &  78.4 & 78.0 & {82.0} & {86.9}&\\
\hdashline[3pt/5pt]

OIM &  0.1 &$\mathrm{KD_{reid}}$&  OIM$_{18}$  & 72.9 & 71.1&80.6 & 85.9\\

OIM &  0.1 &$\mathrm{KD_{reid}}$  &   QEEPS$_{50}$  & {82.4} & {83.0}& {80.8}& {86.0}\\
QEEPS*&  1 & - & -   & 76.6 & 76.0 &- & -\\
QEEPS &  0.1 &$\mathrm{KD_{reid}}$ & OIM$_{50}$   & 82.1 & 81.4& - & - \\
QEEPS &  0.1 &$\mathrm{KD_{reid}}$  & QEEPS$_{50}$& \textbf{84.1} &\textbf{84.3}& - & - \\

\hline
\end{tabular}
\end{center}
\vspace{-0.1cm}

\caption{\footnotesize Knowledge distillation for Resnet18 student models. Above the dashed line, are the detection and person search results of the student model using the two proposed distillation methods ($\mathrm{KD_{det}}$ and $\mathrm{KD_{reid}}$). For $\mathrm{KD_{det}}$, Resnet50 Faster R-CNN detector (DET$_{50}$) is used as teacher. For $\mathrm{KD_{reid}}$ Resnet50 OIM model (OIM$_{50}$) is used as teacher. Below the dashed line, are the results of using  different teachers with OIM and QEEPS as students. OIM$_{18}$ represents Resnet18 OIM model, QEEPS$_{50}$ represents Resnet50 QEEPS model. (*) indicates models trained without KD.}
\label{tab:res18ablation_experiments}
\vspace{-0.2cm}
\end{table}

Next we notice that, our second approach for distillation $\mathrm{KD_{reid}}$ is quite effective also for the model compression scenario. It brings a significant improvement over the baseline on both person search and detection benchmarks. Specifically, it improves by 11.4pp mAP and 12.9pp top-1 for person search and 6.4pp mAP and 4.3pp recall for detection. Quite interestingly, our trained Resnet18 OIM model outperforms Resnet50 baseline OIM model both for person search (78\% mAP \emph{vs} 80.5\%) and detection (75.2\% mAP \emph{vs} 80.6\%). 

Finally, below the dashed line in Table \ref{tab:res18ablation_experiments}, we report results for Resnet18 based OIM and QEEPS student models through supervision from different teacher models such as OIM$_{18}$, OIM$_{50}$, and QEEPS$_{50}$. In particular, supervision of Resnet18 QEEPS with QEEPS$_{50}$ supervision achieves on par performance to the state-of-the-art results of QEEPS$_{50}$ \cite{munjal2019cvpr}.

\section{Conclusions}
\label{sec:conclusions}
We have introduced knowledge distillation for person search and proposed two approaches to supervise either the detector or the re-identification part of two most recent models, OIM~\cite{xiao2017joint} and QEEPS~\cite{munjal2019cvpr}.
In both cases we improve performance on the CUHK-SYSU~\cite{xiao2017joint} and PRW-mini~\cite{munjal2019cvpr} datasets, which extends to model compression.

Our approach is the first-of-a-kind, relaxing the multi-task person search optimization by transferring one task to a teacher. We plan to further investigate whether this multi-task relaxation approach may apply to other multi-task goals.


\bibliography{egbib}
\end{document}